\documentclass[11pt,letterpaper]{article}
\usepackage{emnlp2017}
\usepackage{times}
\usepackage{latexsym}
\usepackage{graphicx,accents}
\usepackage{tikz}
\usepackage{verbatim}
\usetikzlibrary{arrows,shapes}
\usepackage{multirow}

\makeatletter
\DeclareRobustCommand{\cev}[1]{%
	\mathpalette\do@cev{#1}%
}

\newcommand{\do@cev}[2]{%
	\fix@cev{#1}{+}%
	\reflectbox{$\m@th#1\vec{\reflectbox{$\fix@cev{#1}{-}\m@th#1#2\fix@cev{#1}{+}$}}$}%
	\fix@cev{#1}{-}%
}

\newcommand{\fix@cev}[2]{%
	\ifx#1\displaystyle
	\mkern#23mu
	\else
	\ifx#1\textstyle
	\mkern#23mu
	\else
	\ifx#1\scriptstyle
	\mkern#22mu
	\else
	\mkern#22mu
	\fi
	\fi
	\fi
}

\emnlpfinalcopy

\def\emnlppaperid{***}

\title{EmoAtt at EmoInt-2017: Inner attention sentence embedding for Emotion Intensity} 

\author{Edison Marrese-Taylor \and Yutaka Matsuo\\
		Graduate School of Engineering \\
		The University of Tokyo \\
		Tokyo, Japan \\
		 {emarrese,matsuo@weblab.t.u-tokyo.ac.jp}}

\date{}

\begin{document}

\maketitle

\begin{abstract}
	In this paper we describe a deep learning system that has been designed and built for the WASSA 2017 Emotion Intensity Shared Task. We introduce a representation learning approach based on inner attention on top of an RNN. Results show that our model offers good capabilities and is able to successfully identify emotion-bearing words to predict intensity without leveraging on lexicons, obtaining the $13^{th}$ place among 22 shared task competitors. 
\end{abstract}

\section{Introduction}

Twitter is a huge micro-blogging service with more than 500 million tweets per day from different locations in the world and in different languages. This large, continuous, and dynamically updated content is considered a valuable resource for researchers. In particular, many of these messages contain emotional charge, conveying affect—emotions, feelings and attitudes, which can be studied to understand the expression of emotion in text, as well as the social phenomena associated.

While studying emotion in text it is commonly useful to characterize the emotional charge of a passage based on its words. Some words have affect as a core part of their meaning. For example, \textit{dejected} and \textit{wistful} denote some amount of sadness, and are thus associated with sadness. On the other hand, some words are associated with affect even though they do not denote affect. For example, \textit{failure} and \textit{death} describe concepts that are usually accompanied by sadness and thus they denote some amount of sadness.

While analyzing the emotional content in text, mosts tasks are almost always framed as classification tasks, where the intention is to identify one emotion among many for a sentence or passage. However, it is often useful for applications to know the degree to which an emotion is expressed in text. To this end, the WASSA-2017 Shared Task on Emotion Intensity \cite{wassa_emoint_2017} represents the first task where systems have to automatically determine the intensity of emotions in tweets. Concretely, the objective is to given a tweet containing the emotion of joy, sadness, fear or anger, determine the intensity or degree of the emotion felt by the speaker as a real-valued score between zero and one. 

The task is specially challenging since tweets contain informal language, spelling errors and text referring to external content. Given the 140 character limit of tweets, it is also possible to find some phenomena such as the intensive usage of emoticons and of other special Twitter features, such as hashtags and usernames mentions ---used to call or notify other users. In this paper we describe our system designed for the WASSA-2017 Shared Task on Emotion Intensity, which we tackle based on the premise of representation learning without the usage of external information, such as lexicons. In particular, we use a Bi-LSTM model with intra-sentence attention on top of word embeddings to generate a tweet representation that is suitable for emotion intensity. Our results show that our proposed model offers interesting capabilities compared to approaches that do rely on external information sources.

\section{Proposed Approach}

Our work is related to deep learning techniques for emotion recognition in images \cite{Dhall_2015} and videos \cite{EbrahimiKahou_2015}, as well as and emotion classification \cite{lakomkin_2017}. Our work is also related to \newcite{liu_attention-based_2016}, who introduced an attention RNN for slot filling in Natural Language Understanding. Since in the task the input-output alignment is explicit, they investigated how the alignment can be best utilized in encoder-decoder models concluding that the attention mechanisms are helpful. 

EmoAtt is based on a bidirectional RNN that receives an embedded input sequence $x = \{ x_1, ..., x_n \}$ and returns a list of hidden vectors that capture the context each input token $\{ h_1, ..., h_n \}$. To improve the capabilities of the RNN to capture short-term temporal dependencies \cite{mesnil_investigation_2013}, we define the following:

\begin{equation}
\bar{x_i} = [ x_{i-d}; ...; x_{i}; ...; x_{i+d} ]
\end{equation}

Where $\bar{x_i}$ can be regarded as a context window of ordered word embedding vectors around position $i$, with a total size of $2d+1$.  To further complement the context-aware token representations, we concatenate each hidden vector to a vector of binary features $b_i$, extracted from each tweet token, defining an augmented hidden state $\bar h_i = [h_i ; b_i]$. Finally, we combine our $n$ augmented hidden states, compressing them into a single vector, using a global intra-sentence attentional component in a fashion similar to \newcite{vinyals_grammar_2015}. Formally, 
\begin{eqnarray}
u_{j} = v^{\top} \tanh(W_a [\bar h_n;\bar h_j]) \\
\alpha_{j} = \mathrm{softmax}(u_{j}) \\
t = \sum_{j=1}^n \alpha_{j} \cdot \bar h_j
\end{eqnarray}
Where $t$ is the vector that compresses the input sentence $x$, focusing on the relevant parts to estimate emotion intensity. We input this compressed sentence representation into a feed-forward neural network, $\hat y = W_{s} t$, where $\hat y$ is the final predicted emotion intensity. As a loss function we use the mini-batch negative Pearson correlation with the gold-standard.


%
%

\section{Experimental Setup}

To test our model, we experiment using the training, validation and test datasets provided for the shared task \cite{mohammadB17starsem}, which include tweets for four emotions: joy, sadness, fear, and anger. These were annotated using Best-Worst Scaling (BWS) to obtain very reliable scores \cite{kiritchenko-mohammad:2016:N16-11}. 

\begin{table}[h!]
	\footnotesize
	\centering
	\begin{tabular}{c | c | c | c | c}
		\multirow{2}{*}{\textbf{Dataset}} & \multicolumn{3}{|c|}{\textbf{Tweet Length (tokens)}} & \multirow{2}{*}{\textbf{Vocab. in GloVe}}\\
		\cline{2-4}
		& Mean & Min & Max & \\
		\hline
		Fear 	&	17.849 	& 2 	& 37 	& 60.8 \% \\
		Joy 	& 	17.480	& 2		& 42 	& 65.0 \% \\
		Sadness	& 	18.285	& 2		& 38 	& 65.5 \% \\
		Anger	&	17.438	& 1		& 41 	& 65.8 \% \\
		\hline
		\textbf{Average} &   17.776  & 1.75  & 39.5	& 64.3 \%
	\end{tabular}
	\caption{Data summary.}
	\label{table:data_summary}
\end{table}

We experimented with GloVe\footnote{\url{nlp.stanford.edu/projects/glove}} \cite{pennington2014glove} as pre-trained word embedding vectors, for sizes 25, 50 and 100. These are vectors trained on a dataset of 2B tweets, with a total vocabulary of 1.2 M. To pre-process the data, we used Twokenizer \cite{gimpel-EtAl:2011:ACL-HLT2011}, which basically provides a set of curated rules to split the tweets into tokens. We also use Tweeboparser \cite{owoputi-EtAl:2013:NAACL-HLT} to get the POS-tags for each tweet. 

Table \ref{table:data_summary} summarizes the average, maximum and minimum sentence lengths for each dataset after we processed them with Twokenizer. We can see the four corpora offer similar characteristics in terms of length, with a cross dataset maximum length of 41 tokens. We also see there is an important vocabulary gap between the dataset and GloVe, with an average coverage of only 64.3 \%. To tackle this issue, we used a set of binary features derived from POS tags to capture some of the semantics of the words that are not covered by the GloVe embeddings. We also include features for member mentions and hashtags as well as a feature to capture word elongation, based on regular expressions. Word elongation is very common in tweets, and is usually associated to strong sentiment. The following are the POS tag-derived rules we used to generate our binary features.

\begin{itemize}
	\item If the token is an adjective (POS tag = A)
	\item If the token is an interjection (POS tag = !)
	\item If the token is a hashtag (POS tag = \#)
	\item If the token is an emoji (POS tag = E)
	\item If the token is an at-mention, indicating a user as a recipient of a tweet (POS tag = @)
	\item If the token is a verb (POS tag = V)
	\item If the token is a numeral (POS tag = \$)
	\item if the token is a personal pronoun (POS tag = O)

\end{itemize}

\begin{table*}[t]
	\centering
	\begin{tabular}{c | c | c | c | c | c | c | c | c}
		\multirow{2}{*}{\textbf{Corpus}} & \multirow{2}{*}{\textbf{Dropout}} & \multirow{2}{*}{\textbf{Embeddings}} & \multirow{2}{*}{$\lambda$} & \multirow{2}{*}{$h$} & 
		\multicolumn{2}{|c|}{\textbf{EmoAtt}} & \multicolumn{2}{|c}{\textbf{Baseline}}\\
		\cline{6-9}
		& & & & & $\rho_{dev}$ & $\rho_{test}$  & $\rho_{dev}$ &  $\rho_{test}$ \\
		\hline
		Sadness & 0.8     & GloVe Twitter 50 & 0.20 	& 50   & 0.586  & 0.520  & 0.562  & 0.648  \\
		Joy     & 0.8     & GloVe Twitter 50 & 0.20		& 100  & 0.790  & 0.537  & 0.703  & 0.654 \\
		Anger   & 0.5     & GloVe Twitter 50 & 0.01   	& 100  & 0.734  & 0.470  & 0.605  & 0.639 \\
		Fear    & 0.9     & GloVe Twitter 50 & 0.05   	& 100  & 0.644  & 0.561  & 0.574  & 0.652 \\
		\hline
		\multicolumn{5}{c|}{Average} 						   & 0.689  & 0.522  & 0.611  & 0.648
		
	\end{tabular}
	\caption{Summary of the best results.}
	\label{table:result_sumarry}
\end{table*}

While the structure of our introduced model allows us to easily include more linguistic features that could potentially improve our predictive power, such as lexicons, since our focus is to study sentence representation for emotion intensity, we do not experiment adding any additional sources of information as input.

In this paper we also only report results for LSTMs, which outperformed regular RNNs as well as GRUs and a batch normalized version of the LSTM in on preliminary experiments. The hidden size of the attentional component is set to match the size of the augmented hidden vectors on each case. Given this setting, we explored different hyper-parameter configurations, including context window sizes of $1$, $3$ and $5$ as well as RNN hidden state sizes of $100$, $200$ and $300$. We experimented with unidirectional and bidirectional versions of the RNNs.

To avoid over-fitting, we used dropout regularization, experimenting with keep probabilities of $0.5$ and $0.8$. We also added a weighed L2 regularization term to our loss function. We experimented with different values for weight $\lambda$, with a minimum value of 0.01 and a maximum of 0.2.

To evaluate our model, we wrapped the provided scripts for the shared task and calculated the Pearson correlation coefficient and the Spearman rank coefficient with the gold standard in the validation set, as well as the same values over a subset of the same data formed by taking every instance with a gold emotion intensity score greater than or equal to 0.5.

For training, we used mini-batch stochastic gradient descent with a batch size of 16 and padded sequences to a maximum size of 50 tokens, given the nature of the data. We used exponential decay of ratio $0.9$ and early stopping on the validation when there was no improvement after 1000 steps. Our code is available for download on GitHub \footnote{\url{github.com/epochx/emoatt}}.

\section{Results and Discussion}

In this section we report the results of the experiments we performed to test our proposed model. In general, as Table \ref{table:result_sumarry} shows, our intra-sentence attention RNN was able to outperform the Weka baseline \cite{mohammadB17starsem} on the development dataset by a solid margin. Moreover, the model manages to do so without any additional resources, except pre-trained word embeddings. These results are, however, reversed for the test dataset, where our model performs worse than the baseline. This shows that the model is not able to generalize well, which we think is related to the missing semantic information due to the vocabulary gap we observed between the datasets and the GloVe embeddings.

To validate the usefulness of our binary features, we performed an ablation experiment and trained our best models for each corpus without them. Table \ref{table:ablation_study} summarizes our results in terms of Pearson correlation on the development portion of the datasets. As seen, performance decreases in all cases, which shows that indeed these features are critical for performance, allowing the model to better capture the semantics of words missing in GloVe. In this sense, we think the usage of additional features, such as the ones derived from emotion or sentiment lexicons could indeed boost our model capabilities. This is proposed for future work.

\begin{table}[h!]
	\centering
	\begin{tabular}{c | c | c }
		
		\textbf{Dataset} & \textbf{w/features} & \textbf{w/o features} \\
		\hline
		Sadness	& 0.586 & 0.543	\\
		Joy 	& 0.790	& 0.781	\\	
		Anger	& 0.734	& 0.662	\\
		Fear	& 0.644	& 0.561	\\
		\hline
	\end{tabular}
	\caption{Impact of adding our binary features.}
	\label{table:ablation_study}
\end{table}

\begin{figure}[h!]
	\centering
	\includegraphics[width=0.45\textwidth]{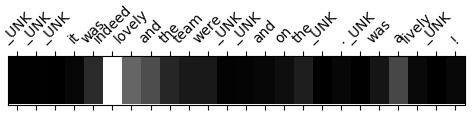}
	\\
	\includegraphics[width=0.45\textwidth]{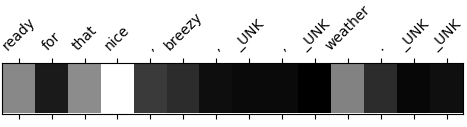}
	\\
	\includegraphics[width=0.45\textwidth]{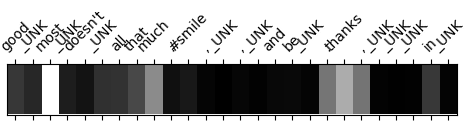}
	\\	\includegraphics[width=0.45\textwidth]{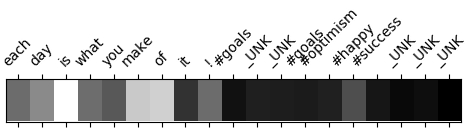}
	\caption{Example of attention weights for the Joy dataset. White denotes more weight.}
	\label{fig:attention}
\end{figure}

On the other hand, our model also offers us very interesting insights on how the learning is performed, since we can inspect the attention weights that the neural network is assigning to each specific token when predicting the emotion intensity. By visualizing these weights we can have a clear notion about the parts of the sentence that the model considers are more important. As Figure \ref{fig:attention} shows, we see the model seems to be have learned to attend the words that naturally bear emotion or sentiment. This is specially patent for the examples extracted from the Joy dataset, where positive words are generally identified. However, we also see some examples where the lack of semantic information about the input words, specially for hashtags or user mentions, makes the model unable to identify some of these the most salient words to predict emotion intensity. Several pre-processing techniques can be implemented to alleviate this problem, which we intend to explore in the future.

\subsection{Anger Dataset}

For the anger dataset, our experiments showed that GloVe embeddings of dimension 50 outperformed others, obtaining an average gain of 0.066 correlation over embeddings of size 25 and of 0.021 for embeddings of size 100. However on ly the first of these values was significant, with a p-value of $3.86 \times 10^{-5}$. Regarding the hidden size of the RNN, we could not find statistical difference across the tested sizes. Dropout also had inconsistent effects, but was generally useful.

\subsection{Joy Dataset}

In the joy dataset, our experiments showed us that  GloVe vectors of dimension 50 again outperformed others, in this case obtaining an average correlation gain of 0.052 ($p = 5.6 \times 10^{-2}$) over embeddings of size 100, and of 0.062 ($p = 3.1 \times 10^{-2}$) for size 25. Regarding the hidden size of the RNN, we observed that 100 hidden units offered better performance in our experiments, with an average absolute gain of 0.052 ($p = 6.5 \times 10^{-2}$) over 50 hidden units. Compared to the models with 200 hidden units, the performance difference was statistically not significant.

\subsection{Fear Dataset}

On the fear dataset, again we observed that embeddings of size 50 provided the best results, offering average gains of 0.12 ($p = 7 \times 10^{-4}$) and 0.11 ($p = 1.9 \times 10^{-3}$) for sizes 25 and 100, respectively. When it comes to the size of the RNN hidden state, our experiments showed that using 100 hidden units offered the best results, with average absolute gains of 0.117 ($p = 9 \times 10^{-4}$) and 0.108 ($p = 0.002.4 \times 10^{-3}$) over sizes 50 and 200.

\subsection{Sadness Dataset}

Finally, on the sadness datasets again we experimentally observed that using embeddings of 50 offered the best results, with a statistically significant average gain of 0.092 correlation points  $(p = 1.3 \times 10^{-3})$ over size 25. Results were statistically equivalent for size 100.  We also observed that using 50 or 100 hidden units for the RNN offered statistically equivalent results, while both of these offered better performance than when using a hidden size of 200.

\section{Conclusions}

In this paper we introduced an intra-sentence attention RNN for the of emotion intensity, which we developed for the WASSA-2017 Shared Task on Emotion Intensity. Our model does not make use of external information except for pre-trained embeddings and is able to outperform the Weka baseline for the development set, but not in the test set. In the shared task, it obtained the $13^{th}$ place among 22 competitors. 

\bibliography{wassa2017emoint}
\bibliographystyle{emnlp_natbib}

\end{document}